\documentclass[conf]{new-aiaa}
\usepackage[utf8]{inputenc}
\usepackage{graphicx}
\usepackage{amsmath}
\usepackage{siunitx}
\usepackage{longtable,tabularx}
\newcommand{\CPA}{{\mathrm{CPA}}}
\newcommand{\Delpred}{{\Delta_\mathrm{pred}}}
\newcommand{\M}{{\mathcal{M}}}
\newcommand{\Nrm}{{\mathcal{N}}}
\newcommand{\Proj}{{\mathcal{P}}}
\newcommand{\Sm}{{\mathcal{S}_m}}
\newcommand{\TCPA}{{T_\mathrm{CPA}}}%

\newcommand{\dvec}{\mathbf{d}}
\newcommand{\dm}{{\dvec_a}}%
\newcommand{\dmo}{{\dvec_o}}%
\newcommand{\dmacpa}{{\dvec^\mathrm{ACPA}_o}}%
\newcommand{\dmapd}{{\dvec^\mathrm{APD}_o}}%
\newcommand{\dvo}{{\dvec^v_o}}%
\newcommand{\dho}{{\dvec^h_o}}%
\newcommand{\rvec}{{\mathbf{r}}}
\newcommand{\mvec}{{\mathbf{m}}}
\newcommand{\tcpa}{{t_\mathrm{CPA}}}%
\newcommand{\ta}{{t_a}}%
\newcommand{\uvec}{{\mathbf{u}}}
\newcommand{\vi}{{\mathbf{v}_i}}
\newcommand{\vo}{{\mathbf{v}_o}}
\newcommand{\hvi}{{\hat{\mathbf{v}}_i}}
\newcommand{\hvo}{{\hat{\mathbf{v}}_o}}
\newcommand{\hvoa}{{\hat{\mathbf{v}}_a}}
\newcommand{\xo}{{\mathbf{x}_o}}

\renewcommand{\xi}{{\mathbf{x}_i}}%

\DeclareMathOperator*{\argmin}{arg\,min}

\title{DAA for an Autonomous Delivery Drone: Maneuver Selection Policies for Near-Collision Risk
  Avoidance}

\author{Franco Callari, Ishay Kamon, Anthony Narkawitz, Pradeep Yarlagadda, and Amir Navot}
\affil{Amazon Prime Air}

\begin{document}

\maketitle

\begin{abstract}
    Detect-and-avoid (DAA) systems for autonomous drones operating in shared airspace must select
    avoidance maneuvers under conditions that differ markedly from those assumed by conventional DAA
    architectures designed for crewed aircraft. Limited on-board sensing capabilities restrict
    detection of collision risk to near distances (sub-km), the time available for decision-making
    is severely limited, uncertainty in the predicted closest point of approach (CPA) and the time
    to reach it is substantial, and the maneuver envelope of a small uncrewed
    aircraft at low altitude is tightly constrained.

    This paper presents and compares two maneuver-selection strategies that address these challenges:
    (1)~a purely geometric approach, steering in the direction away from the intruder at CPA, and (2)~a
    method that also takes into account the known sources of uncertainty in the intruder's location,
    and steers away from a plane expected to contain the intruder's location distribution at CPA.

    Performance is evaluated using the risk ratio metric on a large dataset of simulated near-collision
    encounters with crewed intruders. Results indicate that both strategies are feasible (i.e.,
    computable in closed form from an onboard track estimate), and that the second strategy
    substantially outperforms the first.
\end{abstract}

\section{Introduction}
\label{sec:introduction}

\lettrine{D}{etect} and Avoid (DAA) is a critical enabling capability for the integration of
uncrewed aircraft systems (UAS) into non-segregated airspace~\cite{rtca_do365}. Generally speaking,
UAS operators (especially commercial ones) should follow a multi-layered approach for air traffic
deconfliction, including mission planning, advance re-routing, and coordination with other operators.
However, safe autonomous operation in uncontrolled and traffic-dense environments often requires an
additional independent avoidance layer. This ``last resort'' layer is the context for the present work.

While existing DAA standards such as RTCA DO-365 address cooperative and non-cooperative
surveillance for large remotely-piloted aircraft, the problem takes a fundamentally different
character for slow autonomous drones operating at low altitude in close proximity to other traffic
and terrain. At least four factors distinguish the autonomous drone's DAA problem from the
conventional setting. First, the flight patterns of small aircraft at low altitude are less
predictable (for example, patrol helicopters and hobbyists' paramotors turn more often than
aircraft cruising at high altitude), implying greater uncertainty in predicting conflicts in the
near future. Second, conflicting traffic, which may not be ADS-B equipped, can be reliably detected
and tracked at sub-kilometer ranges only, implying decision times on the order of seconds rather
than tens of seconds. Third, limits in detector performance and short tracking histories imply
significant uncertainties in the estimated intruder state, which propagate into larger uncertainty
in both the predicted CPA geometry and the time remaining before it is reached. Lastly, operational
limits on the drone's speed, turn-rate, and free maneuver space may severely restrict the set of
attainable avoidance maneuvers.

Prior work on UAS DAA has largely focused on well-clear definitions and alerting
logic~\cite{cook2015detecting}, with maneuver guidance often adapted from TCAS-style resolution
advisories~\cite{kochenderfer2012next}. The maneuver-selection problem under the specific
constraints of small autonomous platforms, particularly when the uncertainty envelope around the
predicted CPA is comparable in magnitude to the available maneuver authority, has received
attention only more recently (e.g., in ACAS sXu~\cite{lusk2019acas_sxu}).

This paper addresses the following question: given a detected collision risk with an uncertain
intruder trajectory, how should an autonomous drone select an avoidance maneuver that best
mitigates risk within its constrained maneuver space? We formulate and compare two such maneuver
selection policies:
\begin{itemize}
    \item Away from CPA (ACPA): a simple policy that uses the alert-time estimate of the intruder's location
          at CPA, and steers the ownship directly away from it within the available maneuver envelope.
    \item Away from Plane of Danger (APD): a policy that explicitly attempts to identify at alert time a
          plane in 3D space where collision risk is maximal, and steers the drone away from it.
\end{itemize}
Performance is evaluated using risk ratio metrics applied to a large dataset of simulated
near-collision encounters with crewed intruders. Results indicate that
both strategies are feasible (i.e., computable in closed form from an onboard track estimate),
and that the APD strategy
substantially outperforms ACPA.

\section{Problem Formulation}

\subsection{Maneuver Space}
The extent of a drone's collision-avoidance maneuver will generally be restricted by policy
constraints, e.g., minimum allowable separation between planned drone routes and known ground
obstacles, regulatory limit on ceiling, etc. Further constraints stem from the drone's own
operational environment: maneuvering too far from a planned route at low altitude where many other
drones are operating concurrently can create a secondary conflict risk. Such constraints can be
thought of as defining a maneuver space $\Sm$ surrounding the ownship's planned mission path.
Without much loss of generality, we shall restrict $\Sm$ to be a closed generalized cylinder (a
tube) coaxial to the mission path, which the drone is required to never exit from. Practical and
regulatory considerations will generally limit the cross-section diameter of the maneuver tube to a
few tens of meters, enough to avoid a collision but not so large that it may intersect different
hazards, e.g., ground obstacles or other drones' routes. The cross-section shape will generally be
dictated by a combination of the same constraints and the maneuver authority of the UAS.

\subsection{Avoidance Maneuvers}
\label{sec:maneuver_direction}
The set $\M$ of feasible avoidance maneuvers is bounded by both $\Sm$ and flight control
constraints, e.g., maximum bank angle, climb/descent rate, speed envelope. Therefore, a feasible
maneuver $m \in \M$ is a change of trajectory from the planned mission that keeps the ownship
within $\Sm$, and predictably mitigates the risk of collision. It does so by increasing separation,
that is, by altering the location and time of the predicted CPA toward safer values. Because the
cross section of $\Sm$ is limited to a few tens of meters, the time required for an avoidance
maneuver to attain its maximal allowable deviation in space from the mission path will generally be
on the order of several seconds.

We shall restrict our analysis to \emph{one-shot} avoidance maneuvers, i.e., those where the drone
commits to a pre-programmed maneuver that returns to the original mission path after a few tens of
seconds. Active modification of an already in-progress avoidance maneuver, in response to novel
information about the threat, has limited benefits in our regime of short time available before
CPA, and is out of the scope of this paper.

We shall further restrict the maneuver set $\M$ to purely transverse deviations from the mission
path, i.e., maneuvers that keep unchanged the ownship's velocity along the planned mission path,
and only add a purely transverse, time-varying offset. Such maneuvers consist of three phases:
\begin{itemize}
    \item Initiate: the drone is steered toward the boundary of the maneuver space $\Sm$;
    \item Stay: the drone remains on the boundary of $\Sm$ while flying parallel to the planned mission path,
          for a predefined duration longer than the maximum time-to-CPA the DAA system is designed for;
    \item Return: the drone returns to the mission path.
\end{itemize}

Throughout the maneuver the forward velocity is kept as close as possible to the mission plan. This
restriction is desirable, as it keeps the forward progress of the ownship nearly unchanged, thus
helping with traffic management on commercial UAV routes. However, it may be sub-optimal in cases
where substantially slowing or accelerating the drone may prove beneficial in mitigating the
collision risk.

For simplicity, we shall also assume that the guidance and control laws used in the Initiate and
Return phases are pre-programmed, and parametrized only on the transverse direction followed to
reach the maneuver space boundary.

\begin{figure}[t]
    \centering
    \includegraphics[width=0.55\columnwidth]{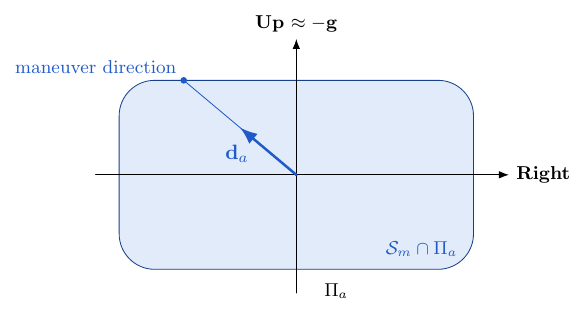}
    \caption{Example of a cross-section $\Sm \cap \Pi_a$ of the maneuver space, and maneuver direction
        vector $\dm$, on the transverse plane $\Pi_a$ at alert time. The drone's forward direction of
        motion is into the page.}
    \label{fig:maneuver_section}
\end{figure}

All such maneuvers can then be uniquely identified by their transverse direction. Specifically, we
will consider maneuvers parametrized by a unit maneuver direction vector $\dm$. The vector is
defined on the plane $\Pi_a$ transverse to the ownship path (i.e., orthogonal to the ownship
velocity) at alert time, see Figure~\ref{fig:maneuver_section} and Section~\ref{sec:kinematics}.

\subsection{The Problem: Maneuver Selection}

Within the above constraints, the problem thus reduces to selecting a maneuver direction that
significantly reduces collision risk at alert time, when the avoidance maneuver is triggered. In
particular, we should expect a well-performing maneuver selection policy to be relatively
insensitive to the uncertainty in the future path of the intruder, given known properties of the
threat detection system.

\subsection{Performance Metric and Evaluation}

Alternative maneuver selection policies may be systematically compared using the risk ratio on a
pre-defined set of encounters. The risk ratio of a maneuver policy $P$ is simply:
\begin{equation}
    \label{eq:risk_ratio}
    R(P) = \frac{C(P)}{C(P_0)}\quad ,
\end{equation}
where $C(\cdot)$ is the count of encounters resulting in a collision or unsafe separation under
policy $P$, and $P_0$ is the null policy where no avoidance maneuver is taken. Unsafe separation
is defined as an event where the distance between ownship and the intruder at CPA is below a
pre-defined threshold. All else being equal, a policy with a \emph{lower} risk ratio is preferred.

For rather obvious safety and practicality reasons, the set of encounters used in the comparison is
created by a realistic simulation of paired trajectories for the UAS and crewed aircraft traffic.
Section~\ref{sec:encounter_data} presents the simulation harness used in this work.

\section{Encounter Kinematics and Notation}
\label{sec:kinematics}

We now formalize the encounter geometry, and introduce the notation used throughout.

\begin{figure}[t]
    \centering
    \includegraphics[width=0.8\columnwidth]{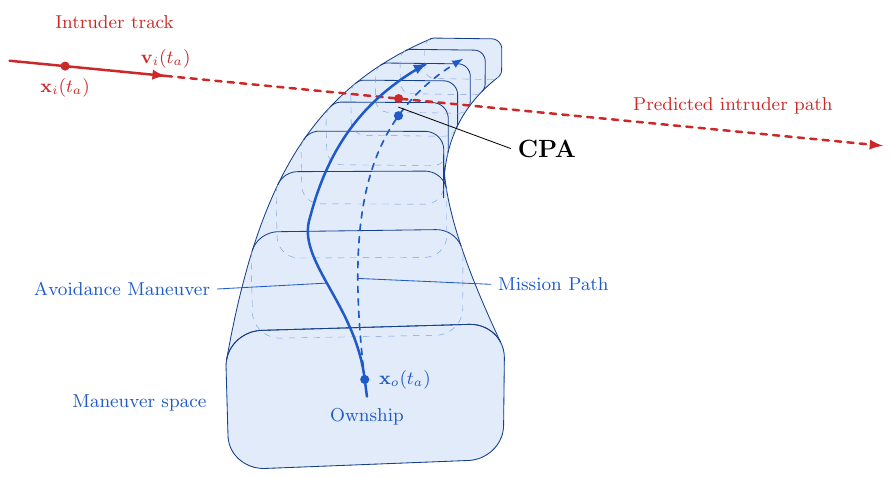}
    \caption{Ownship-intruder encounter and avoidance maneuver.}
    \label{fig:maneuver_tube}
\end{figure}

Consider an autonomous UAS (henceforth ownship, drone) flying along a pre-planned mission path,
while a crewed intruder aircraft flies on a converging path with risk of collision. This is
schematically depicted in Figure~\ref{fig:maneuver_tube}, along with some of the entities defined
below.

We assume that the planned (a.k.a. mission) trajectory of the ownship $\xo(t)$ is precisely known
throughout, along with its velocity $\vo(t)$. At an alert time $\ta$, when an avoidance decision is
to be made, an airborne threat detection system produces an estimate of the intruder state, namely
the intruder's location $\xi(\ta)$ and velocity $\vi(\ta)$. These are input to an intruder motion
model used to predict the future motion of the intruder. Following common practice, we shall use
for this purpose a simple constant velocity assumption, which approximately models an unaware,
non-cooperative intruder:
\begin{equation}
    \xi(t,\, \ta) = \xi(\ta) + \vi(\ta)\cdot (t - \ta)\,,
    \quad t > \ta \quad .
    \label{eq:intruder_model}
\end{equation}

Note that this is a time-varying prediction of the intruder's entire future trajectory, recomputed
as new sensor data is input to the threat detection system, and is affected by uncertainty
depending on the properties of the sensing and tracking technology in use. Ordinarily, it will be
regarded as valid only for a certain duration $T_\mathrm{pred}$, thus defining a prediction
interval $\Delpred$:
\begin{equation}
    \Delpred \triangleq [\ta,\, \ta + T_\mathrm{pred}]\quad.
\end{equation}

To reduce verbosity, in the following we will systematically drop mentions of the alert time $\ta$
when it is obvious from context, for example writing $\xi(t)$ for $\xi(t,\, \ta)$.

Based on this intruder prediction model, we define the predicted line of sight $\rvec(t)$ (a
shorthand for $\rvec(t, \ta)$):
\begin{equation}
    \rvec(t) \triangleq \xi(t) - \xo(t) \, , \quad t \in \Delpred \quad ,
\end{equation}
and the predicted closing distance $r(t)$ from its norm: $r(t) \triangleq \| \rvec(t) \|$. The
predicted time of CPA, $\tcpa$, is the time when the predicted closing distance is
minimized:
\begin{equation}
    \tcpa \triangleq \argmin_{t \in \Delpred} r(t)\quad .
\end{equation}
The predicted CPA is then simply defined as the pair of ownship and intruder locations at the
predicted time of CPA:
\begin{equation}
    \CPA \triangleq \{\xo(\tcpa),\, \xi(\tcpa)\} \quad ,
\end{equation}
and the predicted time-to-CPA as the amount of time left before the CPA is attained:
\begin{equation}
    \TCPA \triangleq \tcpa - \ta\quad .
\end{equation}
The predicted miss vector $\mvec$ is the displacement from ownship to intruder at the CPA:
\begin{equation}
    \mvec \triangleq \rvec(\tcpa) = \xi(\tcpa) - \xo(\tcpa)\quad ,
\end{equation}
and its length $\|\mvec\|$ is the predicted miss distance. Methods for estimating the time-to-CPA
from images have previously been investigated in~\cite{glozman2021vision}.

Unless otherwise noted, vectors are decomposed in the local North-East-Down (NED) frame at the
location of interest. The gravity direction vector is then denoted with $\mathbf{g} = (0, 0, 1)^T$,
and we will systematically assume it is the same at both alert location and predicted CPA. We make
use of the normalization operator $\Nrm$ for a non-null vector $\uvec$:
\begin{equation}
    \Nrm(\uvec) \triangleq \uvec \, / \, \|\uvec\| \quad ,
\end{equation}
and denote the unit vectors of the ownship velocity at alert and CPA times, and of the predicted
intruder velocity (constant, by Equation~\eqref{eq:intruder_model}), respectively as:
\begin{equation}
    \hvoa \triangleq \Nrm\big(\vo(\ta)\big)\,,\qquad
    \hvo \triangleq \Nrm\big(\vo(\tcpa)\big)\,,\qquad
    \hvi \triangleq \Nrm\big(\vi(\ta)\big) = \Nrm\big(\vi(\tcpa)\big)\,.
\end{equation}
We will make use of the projection operators onto the transverse plane at alert $\Pi_a$, and at CPA
$\Pi_o$, defined for a non-null vector $\uvec$ respectively as:
\begin{equation}
    \Proj_a(\uvec) \triangleq \uvec - (\uvec \cdot \hvoa)\, \hvoa, \qquad
    \Proj_o(\uvec) \triangleq \uvec - (\uvec \cdot \hvo)\, \hvo
    \quad .
\end{equation}

As we are mainly interested in collision avoidance when the ownship is in cruise flight, we assume
that $\|\vo\|$ is bounded away from zero, and that the ownship pitch is moderate at both alert and
CPA, i.e., $\|\hvoa \times \mathbf{g}\| > 0$ and $\|\hvo \times \mathbf{g}\| > 0$. Planes $\Pi_a$
and $\Pi_o$ are then both well defined, and so are the projections of the gravity vector onto them.
We shall also use the sign function $\mathrm{sgn}(x)$, equal to $1$ for $x > 0$ and to $-1$
otherwise.

Again, we remark that all of the above quantities depend on the instant at which the intruder state
is estimated, and are recomputed as new sensor data arrives.

\section{Maneuver Selection Policies}
\label{sec:policies}

\begin{figure}[t]
    \centering
    \includegraphics[width=0.82\columnwidth]{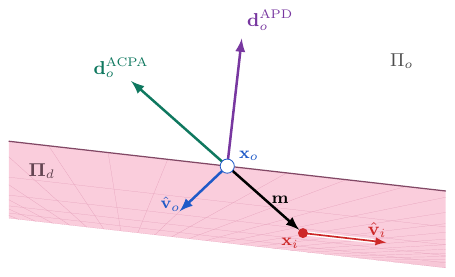}
    \caption{A sketch of the maneuver directions selected by the two policies discussed
        in this work. The page represents the transverse plane at CPA, $\Pi_o$, observed from behind
        the drone. The drone itself is at $\xo$, flying into the page along $\hvo$.
        The intruder is predicted to pass at $\xi$, the tip of the miss vector $\mvec$, traveling
        along $\hvi$. Policy ACPA steers the drone opposite to the miss vector. The \emph{Plane of Danger},
        $\Pi_d$, spanned by $\hvi$ and $\hvo$, is the pink band shown receding into the page.
        Policy APD steers the drone normal to $\Pi_d$, toward the side away from the intruder.}
    \label{fig:policy_plane}
\end{figure}

We now formalize the two maneuver selection policies briefly introduced in
Section~\ref{sec:introduction}. Both are computed in closed-form at alert time, and produce the
unit commanded maneuver direction vector $\dm$ of Section~\ref{sec:maneuver_direction}. The
maneuvers the two policies prescribe for a single encounter are illustrated in
Figure~\ref{fig:policy_plane}.

In the following we will use a subscript to denote the transverse plane a vector lies in. Thus the
commanded maneuver direction $\dm$ lies in $\Pi_a$. We will use a superscript on $\dvec$ to
indicate the policy in use: $\dvec^\mathrm{ACPA}$ or $\dvec^\mathrm{APD}$. Finally, we will use the
superscripts $v$ or $h$ on $\dvec$ to denote vertical and horizontal fallback directions, used in
edge cases where the maneuver direction would otherwise be undetermined. Specifically, the three
fallback directions of interest are:
\begin{equation}
    \dvo \triangleq \Nrm\big(\Proj_o(\mathbf{g})\big) \quad , \qquad
    \dho \triangleq \Nrm\big(\dvo \times \hvo\big) \quad , \qquad
    \dvec^v_a \triangleq \Nrm\big(\Proj_a(\mathbf{g})\big) \quad .
\end{equation}

The computation of both policies proceeds in two steps. The first step is the policy proper. It
yields an intermediate direction $\dmo$, a unit vector lying in the plane $\Pi_o$ transverse to the
mission-planned ownship velocity at the predicted CPA. That is the plane in which both policies
reason, as the geometrical entities they rely on, the miss vector and the plane of danger, are
defined at CPA.

The second step, common to both policies, delivers the actual maneuver direction $\dm$ as the
normalized projection of $\dmo$ onto $\Pi_a$, falling back on the vertical direction $\dvec^v_a$
when the projection is too short:
\begin{equation}
    \dm =
    \begin{cases}
        \Nrm\big(\Proj_a(\dmo)\big) & \text{if }\,\, \|\Proj_a(\dmo)\| \ge \theta_p \\
        \dvec^v_a                   & \text{otherwise}
    \end{cases} \quad .
    \label{eq:alert_plane_projection}
\end{equation}
The threshold $\theta_p$, set to 0.3 in our experiments (see Table~\ref{tab:pmd_thresholds}),
guards against a $\dmo$ nearly parallel to the ownship velocity at alert: the direction of such a
$\dmo$ within $\Pi_a$ would be dictated by estimation noise. This fallback is triggered only when
the ownship turns sharply between alert and predicted CPA, by more than \ang{72.5} for the above
value of $\theta_p$, which is sharper than would ordinarily be observed in a cruise encounter of
short time-to-CPA; see Appendix~\ref{app:projection_bound} for a proof. Note also that, outside
this degenerate case, the second step carries over the content of $\dmo$ within $\Pi_a$ exactly:
the projection removes only its component along $\hvoa$, and the renormalization merely rescales
what remains. The whole effect of the second step is thus the test against $\theta_p$.

\subsection{Maneuver Away from CPA}

The restriction to maneuvers that leave the along-path velocity unaffected suggests an intuitive
maneuver policy, namely to steer directly away from where the intruder is predicted to pass. More
precisely:
\begin{quote}
    \textbf{ACPA Policy:} \emph{Steer the ownship opposite to the transverse component of the miss vector.}
\end{quote}

The ACPA direction is thus opposite to the component of the miss vector on $\Pi_o$:
\begin{equation}
    \label{eq:acpa_general}
    \dmacpa =
    \begin{cases}
        -\Nrm\big(\Proj_o(\mvec)\big) & \text{if }\,\, \|\Proj_o(\mvec)\| \ge \theta_m \\
        \dvo                          & \text{otherwise}
    \end{cases} \quad ,
\end{equation}
where the threshold $\theta_m$ guards against the case of an intruder predicted to pass nearly
through the ownship's own location at CPA, when the pass direction cannot be reliably estimated.

Note that the policy neglects the change in location and time of CPA induced by the maneuver
itself, an approximation that relies on the intruder being unaware and flying a mostly rectilinear
path at near-constant speed.

\subsection{Maneuver Away from the Plane of Danger}

This second policy attempts to account for how the uncertainty in the intruder location at alert
time affects the CPA estimate. Specifically, we consider the case where errors in the 3D location
of the intruder's track are dominated at all times by their radial component with respect to the
ownship. This is a common situation, for example, when the ownship's sensors are ordinary cameras
and the threat detection method is based on computer vision, as the \emph{direction} from ownship
to intruder can be accurately estimated from the ownship's navigation state and the optical and
mechanical calibration of the cameras, but the intruder's \emph{distance} can be bracketed with
much lower precision, e.g., by a stereo-based algorithm.\footnote{Note that the opposite can be
    expected to occur when the ownship uses a ranging sensor to detect the intruder. For example, a
    Doppler radar will ordinarily produce accurate distance and velocity estimates with low angular
    resolution.}

For an intruder cruising on a mostly rectilinear trajectory, large range errors place the errors in
its estimated position along the line of sight $\rvec = \xi - \xo$. The intruder velocity,
estimated from successive positions, then has errors mostly along $\rvec$ as well, so the estimated
$\vi$ lies approximately in the plane spanned by $\rvec$ and the true intruder velocity. Under
constant velocities, the line of sight forms an angle $\beta$ with the relative velocity $\vi -
    \vo$, with $\sin\beta = \|\mvec\| / \|\rvec\|$. At alert time, when $\|\rvec\| \gg \|\mvec\|$, the
two are therefore closely aligned, and that plane, spanned by $\rvec$ and the true intruder
velocity, approximately contains both $\vi$ and $\vo$.\footnote{This approximation is exact when
    the trajectories of ownship and intruder are rectilinear with constant velocity, and degrades
    gracefully when either aircraft turns.}

The predicted intruder location at CPA, $\xi(\tcpa) = \xi(\ta) + \vi\, \TCPA$, inherits the same
error structure, so the intruder locations at CPA are plausibly clustered near the plane spanned by
$\vi$ and $\vo$ at CPA. We call it the \emph{Plane of Danger}, $\Pi_d$.

A formal estimation of $\Pi_d$ would require propagating the range error through detection and
tracking, then extracting the first two principal components of the resulting error distribution.
Here we instead make the simple approximation that $\Pi_d$ is the plane spanned by the estimated
intruder velocity $\vi$ and the planned ownship velocity $\vo$ at the predicted CPA. Intuitively,
we should then steer the ownship away from the plane of danger, on the side opposite the predicted
intruder location. Because we restrict ourselves to policies that only modify the ownship
transverse velocity, we must choose a direction in $\Pi_o$. Among those directions, the one
departing fastest from the plane of danger is its normal, which already lies in $\Pi_o$, being
orthogonal to $\hvo$. Hence the prescription:
\begin{quote}
    \textbf{APD Policy:} \emph{Steer the ownship in the direction of the cross product of the planned
        ownship velocity at the predicted CPA and the estimated intruder velocity, oriented in
        opposition to the miss vector.}
\end{quote}

Let $\mathbf{n} \triangleq \Nrm(\hvo \times \hvi)$ be the normal to $\Pi_d$. Of the two sides of
$\Pi_d$, the ownship should maneuver away from the one containing the tip of the miss vector,
hence:
\begin{equation}
    \dmapd = -\mathrm{sgn}(\mvec \cdot \mathbf{n})\,\mathbf{n} \quad .
    \label{eq:apd_general}
\end{equation}
Equation~\eqref{eq:apd_general} is undefined or numerically unstable in three situations: a
quasi-stationary intruder, nearly parallel velocities, and a miss vector close to the plane of
danger. The treatment of these special cases, given in Appendix~\ref{app:apd_degenerate}, completes
the definition of the policy.

This policy, too, neglects the changes in the time and location of the CPA due to the maneuver.

\section{Evaluation Methodology}

\subsection{Encounter Data}
\label{sec:encounter_data}
The performance of each proposed maneuver policy is evaluated on a dataset consisting of
\num{10000} pairs of simulated encounters between a commercial delivery drone and aircraft
of four types, namely helicopters, fixed wing aircraft, ultralights and hot-air balloons.
The two encounters in each pair share the same intruder trajectory and planned mission path for
the drone. They differ only in the maneuver policy being used, i.e., the Null policy (no maneuver)
in one, or the policy under test (ACPA or APD) in the other.

The simulated ownship is an exact model of the flight controller of an Amazon Prime Air MK30
commercial delivery drone, flying operationally representative missions. The simulated trajectories
of the three non-balloon types are Monte Carlo--generated from MIT-LL uncorrelated models trained
on VFR non-participating traffic~\cite{underhill2021uncorrelated}, using the implementation
in~\cite{weinert2021manned_bayes}. The model samples are randomly rotated in heading, then
translated to create encounters with the drone. The hot-air balloon trajectories are instead
generated by a purpose-built model, as the corresponding published MIT-LL model does not
parametrize the accelerations and turn rates our simulator requires. The repeatable, pseudo-random
intruder trajectories span a range of geometries, closure rates, and the four intruder types above.

Our simulation harness processes at \SI{10}{\hertz} simulated cycle time, flying both ownship and
intruder along the trajectories of a sample encounter, with the intruder locations being input to a
model of the drone's on-board airborne threat detection software. Details of the detector model
itself are out of scope for the present work: suffices to say that it produces, with probability
equal to the distance-dependent detection rate of the real onboard threat detection software,
reports containing an estimated point location in 3D space for the detected object. The errors in
the 3D location faithfully reproduce the statistics of the onboard system.

Use of a detector model allows us to efficiently run thousands of simulations without incurring the
complexity and computational cost of directly simulating the onboard sensors. Instead, using the
same software as the real onboard system, the detector model's per-cycle reports are tracked over
time in 3D position and velocity, and the CPA of each track is estimated with the usual
constant-velocity assumption. Finally, each track is classified as being a collision threat or not,
producing an alert/clear binary output.

When the software outputs an alert in the encounter with non-null maneuver policy, the
policy-selected avoidance maneuver is commanded and run with exactly the same control laws as the
real drone. The one-shot avoidance maneuver ramps off the mission path along the commanded
transverse maneuver direction, levels off at the edge of the maneuver space keeping the mission
along-path velocity, then returns to the mission path with a similar ramp \SI{20}{\second} after it
is initiated. In the experiments reported in this paper both maneuver policies use a maneuver space
with circular cross-section of \SI{30}{\meter} radius.

Additionally, the simulation harness uses a simple elevation-dependent wind model with fixed
direction. The fixed direction is not too unrealistic an approximation, given the short range of
distances and time duration of the encounters from first detection to CPA: less than
\SI{1}{\kilo\meter} and \SI{20}{\second} respectively.

\subsection{Policy Computation}
\label{sec:policy_computation}
Both policies require only a handful of vector dot and cross products per alert, so their realtime
computational cost is constant and ordinarily negligible, compared with that of intruder detection
and tracking.

The threshold values used in the policy calculations for all experiments in this paper are listed
in Table~\ref{tab:pmd_thresholds}.
\begin{table}[htb]
    \centering
    \caption{Thresholds used in the computation of the maneuver selection policies.}
    \label{tab:pmd_thresholds}
    \begin{tabular}{lll}
        \hline
        Symbol          & Meaning                                                      & Value                       \\
        \hline
        $\theta_v$      & Intruder speed below which the intruder is deemed stationary & \SI{3.0}{\meter\per\second} \\
        $\theta_\alpha$ & Angle below which velocities are deemed (anti-)parallel      & \SI{3}{\degree}             \\
        $\theta_m$      & Minimum miss-vector component for a reliable side estimate   & \SI{1}{\meter}              \\
        $\theta_g$      & Minimum vertical component of $\mathbf{n}$                   & $10^{-3}$                   \\
        $\theta_p$      & Minimum norm of the projection of $\dmo$ onto $\Pi_a$        & \num{0.3}                   \\
        \hline
    \end{tabular}
\end{table}

\subsection{Encounter Scoring}

An encounter is deemed to have a collision event at a given time according to the following
criteria:
\begin{itemize}
    \item For helicopter, fixed wing and ultralight intruders, whenever they penetrate a vertical cylinder
          centered on the ownship, with radius \SI{7}{\meter} and full height \SI{7}{\meter}, i.e., extending
          \SI{3.5}{\meter} above and below the ownship.
    \item For hot-air balloons, whenever they penetrate a sphere centered on the ownship, with radius equal
          to \SI{17.5}{\meter}.
\end{itemize}

\subsection{Metrics Aggregation}
Each simulated encounter is scored with a single boolean outcome: whether the above collision
criteria were met at any simulated cycle of the encounter. The two members of a pair are then
cross-tabulated, yielding four counts of encounters:
\begin{itemize}
    \item $n_{oc}$: The number of encounters where a collision occurred under the null policy, but
          did not occur under the evaluated policy. These are the encounters where the active
          policy \emph{avoided} a collision.
    \item $n_{co}$: The number of encounters where no collision occurred under the null policy, but
          did occur under the evaluated policy. Encounters where the active policy
          \emph{caused} a collision.
    \item $n_{cc}$: The number of encounters where a collision occurred under both the null and
          the active policy. Encounters where the active policy made \emph{no difference}, the
          outcome being unsafe either way.
    \item $n_{oo}$: The number of encounters where no collision occurred under either the null
          policy or the one being evaluated. Encounters where the active policy made \emph{no
              difference}, the outcome being safe either way.
\end{itemize}

The count function $C(\cdot)$ in Equation~\eqref{eq:risk_ratio} is then simply given by:
\begin{align}
    C(P)   & = n_{co} + n_{cc} \quad , \\
    C(P_0) & = n_{oc} + n_{cc} \quad .
\end{align}
The counts are unweighted: every encounter of the dataset contributes equally, and no attempt is
made to reweight the intruder types by their expected frequency of occurrence in the airspace.
Risk ratios are therefore reported both overall and per aircraft type, the latter being the
quantity that transfers to an airspace with a different traffic mix.

\section{Results}

The counts and risk ratios for the ACPA and APD maneuver policies, overall and by intruder type,
are summarized in Table~\ref{tab:risk_ratio_30m}. The last column, $R_{95}$, shows the one-sided
\SI{95}{\percent} upper confidence limit on $R(P)$, obtained by treating $C(P)$ as Poisson and
conditioning on the observed $C(P_0)$. The APD policy clearly dominates: apart from Balloons, where
the two policies behave identically, every ACPA risk ratio lies above the corresponding $R_{95}$
value of APD.

The table also shows that the simulation harness produces exactly matched null-policy encounters.
Both policies are evaluated on the same encounter pairs, with the same pseudorandom draws for the
simulated detector model, and under the null policy the ownship flies its mission path unperturbed.
Therefore $C(P_0)$ ought not to depend on the policy at all, and we experimentally verify that it
does not: the null-policy counts agree exactly between the two policies, overall and for every
aircraft type. The $n_{oo}$ values do differ between the policies, but only by $n_{co}$, the number
of collisions each policy caused, which is a property of the policy and not of the harness.

\begin{table}[htb]
    \centering
    \caption{Evaluation counts, risk ratio, and its one-sided \SI{95}{\percent} upper confidence
        limit $R_{95}$, by maneuver policy and intruder type.}
    \label{tab:risk_ratio_30m}
    \small
    \sisetup{detect-weight = true, detect-family = true}
    \begin{tabular}{llrrrrrrrrr}
        \hline
        Policy       & Intruder Type & Enc. Pairs           & $n_{oc}$     & $n_{co}$   & $n_{cc}$    & $n_{oo}$      & $C(P)$      & $C(P_0)$     & $R(P)$ (\%)   & $R_{95}$ (\%) \\
        \hline
        ACPA         & All           & \num{10000}          & 947          & 48         & 33          & 8972          & 81          & 980          & 8.27          & 9.94          \\
                     & Balloons      & \num{1339}           & 565          & 0          & 1           & 773           & 1           & 566          & 0.18          & 0.84          \\
                     & Fixed wing    & \num{2679}           & 105          & 9          & 13          & 2552          & 22          & 118          & 18.64         & 26.62         \\
                     & Helicopters   & \num{4585}           & 218          & 22         & 16          & 4329          & 38          & 234          & 16.24         & 21.29         \\
                     & Ultralight    & \num{1397}           & 59           & 17         & 3           & 1318          & 20          & 62           & 32.26         & 46.87         \\
        \hline
        \textbf{APD} & \textbf{All}  & \textbf{\num{10000}} & \textbf{955} & \textbf{9} & \textbf{25} & \textbf{9011} & \textbf{34} & \textbf{980} & \textbf{3.47} & \textbf{4.62} \\
                     & Balloons      & \num{1339}           & 565          & 0          & 1           & 773           & 1           & 566          & 0.18          & 0.84          \\
                     & Fixed wing    & \num{2679}           & 107          & 1          & 11          & 2560          & 12          & 118          & 10.17         & 16.48         \\
                     & Helicopters   & \num{4585}           & 223          & 5          & 11          & 4346          & 16          & 234          & 6.84          & 10.39         \\
                     & Ultralight    & \num{1397}           & 60           & 3          & 2           & 1332          & 5           & 62           & 8.06          & 16.96         \\
        \hline
    \end{tabular}
\end{table}

The dominance of APD is further illustrated by McNemar's exact test~\cite{mcnemar1947} on the
outcomes of the two policies over matched encounters, whose encounter counts and two-sided
$p$-values are summarized in Table~\ref{tab:mcnemar}. This is the appropriate test for such data,
as it conditions on the discordant encounters, namely those in which exactly one of the two
policies resulted in a collision. We conclude that policy APD is favoured at $p < 0.01$ for the
three non-balloon intruder types, so the Bonferroni correction~\cite{bonferroni1936,dunn1961} with
threshold $0.05/3 = 0.0167$ does not change the conclusion. The encounters with balloons admit no
test, as neither policy collided in any of them in which the other did not.

\begin{table}[htb]
    \centering
    \caption{McNemar's exact test of ACPA against APD on matched encounters.
        The discordant counts $b$ and $c$ are the encounters in which only ACPA, and
        only APD, resulted in a collision.}
    \label{tab:mcnemar}
    \small
    \sisetup{detect-weight = true, detect-family = true}
    \begin{tabular}{lrrrrrll}
        \hline
        Intruder Type & Enc. Pairs           & Both        & $b$ (ACPA only) & $c$ (APD only) & Neither             & $p$                   & Favours      \\
        \hline
        \textbf{All}  & \textbf{\num{10000}} & \textbf{25} & \textbf{56}     & \textbf{9}     & \textbf{\num{9910}} & \textbf{\num{2.0e-9}} & \textbf{APD} \\
        Balloons      & \num{1339}           & 1           & 0               & 0              & \num{1338}          & ---                   & tie          \\
        Fixed wing    & \num{2679}           & 11          & 11              & 1              & \num{2656}          & \num{6.3e-3}          & APD          \\
        Helicopters   & \num{4585}           & 11          & 27              & 5              & \num{4542}          & \num{1.1e-4}          & APD          \\
        Ultralight    & \num{1397}           & 2           & 18              & 3              & \num{1374}          & \num{1.5e-3}          & APD          \\
        \hline
    \end{tabular}
\end{table}

\section{Discussion and Future Work}
\label{sec:discussion}

Perhaps surprisingly, the policy of steering away from the plane of danger achieves substantially
better results than the intuitive one of simply steering directly away from the danger itself, as
represented by the best estimate (at alert time) of the intruder location at CPA. This indicates
that careful consideration of the sources of error in the CPA estimates is necessary for successful
maneuver selection.

The above results use a maneuver space with circular cross-section of \SI{30}{\meter} radius, which
presumes comparable control authority and routing constraints in every transverse direction
throughout a mission. The ordering of the two policies is not invariant to that choice. On the same
encounters, but using a maneuver space with elliptical cross-section, still with \SI{30}{\meter}
lateral, but \SI{15}{\meter} vertical semi-axes, the ordering of the policies reverses, with ACPA
at a risk ratio of \SI{8.8}{\percent} against \SI{12.7}{\percent} for APD. The reason is that APD
selects a mostly-vertical direction in the large majority of these encounters, so halving the
vertical extent halves the displacement it actually achieves, whereas ACPA spreads its directions
more evenly around the transverse plane. A policy intended for an anisotropic maneuver space should
thus select the direction accounting for the attainable displacement in each direction, rather than
based on geometry alone; this is a natural extension of the present work.

Both policies described in this work essentially ignore the effect of the kinematics of the
avoidance maneuver itself in relation to the intruder's path, and the resulting changes in the
location and time of the CPA compared to the non-maneuvering ownship path. These issues will be the
subject of future work.

\appendix
\section{Degenerate Cases of Policy APD}
\label{app:apd_degenerate}

The three situations in which Equation~\eqref{eq:apd_general} is undefined or numerically unstable
are tested in the order below. The first condition that applies determines the chosen maneuver
direction, and the general case of Equation~\eqref{eq:apd_general} is used only if none of them
does. The threshold values are those of Table~\ref{tab:pmd_thresholds}.

\paragraph{APD special case 1: stationary intruder.}
When the intruder velocity is small, $\|\vi\| < \theta_v$, the direction of the intruder velocity
is dominated by estimation noise, and so is the plane of danger. However, for a (nearly) stationary
intruder, the radial location errors are spread along the line of sight from ownship to intruder,
which lies approximately in the plane spanned by $\hvo$ and $\mvec$. We therefore take this plane
as the plane of danger, and maneuver toward its downward side:
\begin{equation}
    \dmapd =
    \begin{cases}
        \mathrm{sgn}(\mathbf{n}_s \cdot \mathbf{g})\, \Nrm(\mathbf{n}_s) & \text{if } \|\Proj_o(\mvec)\| \ge \theta_m \\
        \dho                                                             & \text{otherwise}
    \end{cases}
    \quad , \qquad \mathbf{n}_s \triangleq \hvo \times \mvec \quad .
\end{equation}
Note that $\|\mathbf{n}_s\| = \|\Proj_o(\mvec)\|$, so that the threshold test also ensures that
$\mathbf{n}_s$ is well defined. In this case the horizontal default is preferred, because
near-stationary intruders may actually be ground-fixed obstacles (e.g., the tips of cell towers)
misidentified as airborne ones.

\paragraph{APD special case 2: parallel velocities.}
When the ownship and intruder velocities are nearly parallel or anti-parallel, $\|\hvo \times
    \hvi\| < \sin\theta_\alpha$, and the plane of danger is again poorly defined. We replace it with
the plane containing $\hvo$ and the horizontal axis orthogonal to it, whose normal in $\Pi_o$ is
$\dvo$. The ownship then maneuvers vertically toward the half-space that does not contain the
intruder at CPA:
\begin{equation}
    \dmapd =
    \begin{cases}
        -\mathrm{sgn}(\mvec \cdot \dvo)\, \dvo & \text{if } |\mvec \cdot \dvo| \ge \theta_m \\
        \dvo                                   & \text{otherwise}
    \end{cases} \quad .
\end{equation}

\paragraph{APD special case 3: miss vector close to the plane of danger.}
When $|\mvec \cdot \mathbf{n}| < \theta_m$, the predicted intruder location at CPA lies too close
to $\Pi_d$ for the sign in Equation~\eqref{eq:apd_general} to be reliable. Consistent with the
downward default, the ownship then maneuvers toward the side of $\Pi_d$ facing the ground, unless
$\Pi_d$ is itself nearly vertical, in which case the orientation of $\mathbf{n}$ given by the cross
product is retained:
\begin{equation}
    \dmapd =
    \begin{cases}
        \mathrm{sgn}(\mathbf{n} \cdot \mathbf{g})\, \mathbf{n} & \text{if } |\mathbf{n} \cdot \mathbf{g}| \ge \theta_g \\
        \mathbf{n}                                             & \text{otherwise}
    \end{cases} \quad .
\end{equation}

\section{Reachability of the Projection Fallback}
\label{app:projection_bound}

The fallback of Equation~\eqref{eq:alert_plane_projection} is taken when the projection of $\dmo$
onto $\Pi_a$ is shorter than $\theta_p$. Here we show that this requires a large turn of the
ownship velocity between alert and predicted CPA.

Let $\delta$ be the angle between $\hvoa$ and $\hvo$, and decompose $\hvoa = \hvo\cos\delta +
    \mathbf{u}\sin\delta$ for some unit vector $\mathbf{u}$ in $\Pi_o$. Since $\dmo \cdot \hvo = 0$ by
construction, and $\|\dmo\| = \|\mathbf{u}\| = 1$:
\begin{equation}
    |\dmo \cdot \hvoa| = |\sin\delta| \, |\dmo \cdot \mathbf{u}| \le |\sin\delta| \quad ,
\end{equation}
and therefore:
\begin{equation}
    \|\Proj_a(\dmo)\|^2 = 1 - (\dmo \cdot \hvoa)^2 \ge 1 - \sin^2\delta = \cos^2\delta \quad .
    \label{eq:projection_bound}
\end{equation}
The fallback thus requires $|\cos\delta| \le \|\Proj_a(\dmo)\| < \theta_p$, which for $\theta_p =
    0.3$ is only possible when $\delta > \arccos\theta_p \approx \ang{72.5}$.

\bibliography{references}

\end{document}